\documentclass[conference]{IEEEtran}
\IEEEoverridecommandlockouts
\usepackage{cite}
\usepackage{amsmath}
\usepackage{amssymb}
\usepackage{graphicx}
\usepackage{hyperref}
\usepackage{caption}
\title{AI-Driven Three-Dimensional Reconstruction and Quantitative Analysis for Burn Injury Assessment}

\author{\IEEEauthorblockN{S. Kalaycioglu\IEEEauthorrefmark{1}\IEEEauthorrefmark{2}\IEEEauthorrefmark{3}\IEEEauthorrefmark{4}, C. Hong\IEEEauthorrefmark{5}\IEEEauthorrefmark{6}, K. Zhai\IEEEauthorrefmark{4},  H. Xie\IEEEauthorrefmark{3}\IEEEauthorrefmark{4}, J.N. Wong\IEEEauthorrefmark{7}}
\IEEEauthorblockA{\IEEEauthorrefmark{1}Toronto Metropolitan University\\
\IEEEauthorrefmark{2}York University\\
\IEEEauthorrefmark{3}AIMechatroniX Inc.\\
\IEEEauthorrefmark{4}DrRobot Inc.\\
\IEEEauthorrefmark{5}Skinopathy Inc.\\
\IEEEauthorrefmark{6}Scarborough Health Network, Centenary Hospital, Scarborough, Ontario, Canada\\
\IEEEauthorrefmark{7}University of Alberta Hospital, Edmonton, Alberta, Canada}
}

\begin{document}

\maketitle

\begin{abstract}
Accurate, reproducible burn assessment is critical for treatment planning, healing monitoring, and medico-legal documentation, yet conventional visual inspection and 2D photography are subjective and limited for longitudinal comparison. This paper presents an AI-enabled burn assessment and management platform that integrates multi-view photogrammetry, 3D surface reconstruction, and deep learning–based segmentation within a structured clinical workflow. Using standard multi-angle images from consumer-grade cameras, the system reconstructs patient-specific 3D burn surfaces and maps burn regions onto anatomy to compute objective metrics in real-world units, including surface area, TBSA, depth-related geometric proxies, and volumetric change. Successive reconstructions are spatially aligned to quantify healing progression over time, enabling objective tracking of wound contraction and depth reduction. The platform also supports structured patient intake, guided image capture, 3D analysis and visualization, treatment recommendations, and automated report generation. Simulation-based evaluation demonstrates stable reconstructions, consistent metric computation, and clinically plausible longitudinal trends, supporting a scalable, non-invasive approach to objective, geometry-aware burn assessment and decision support in acute and outpatient care.
\end{abstract}

\begin{IEEEkeywords}
Burn wound assessment, three-dimensional reconstruction, photogrammetry, artificial intelligence, deep learning, computer vision, Structure from Motion, TBSA estimation, longitudinal healing analysis, clinical decision support
\end{IEEEkeywords}

\section{Introduction}
Burn injuries represent a significant global health burden, accounting for substantial mortality, morbidity, and long-term disability worldwide. According to the World Health Organization, approximately 180,000 burn-related deaths occur annually, with a disproportionate impact in low- and middle-income regions where access to specialized burn care and longitudinal follow-up is limited \cite{who2023}. Even in well-resourced healthcare systems, burns remain among the most complex injuries to assess and manage due to their dynamic evolution, heterogeneous depth, and profound systemic effects. Accurate early assessment and consistent longitudinal documentation are therefore critical for guiding resuscitation, surgical decision-making, referral to burn centers, and evaluation of healing outcomes.

Despite advances in burn surgery, critical care, and infection control, the assessment and documentation of burn injuries remain largely subjective and manual. Burn size estimation, typically expressed as percentage total body surface area (\%TBSA), and burn depth classification directly influence fluid resuscitation strategies, operative planning, and prognosis \cite{wallace1951, lund1944, hettiaratchy2004a, hettiaratchy2004b}. However, widely used tools such as the Rule of Nines and the Lund–Browder charts are dependent on clinician experience and visual judgment, resulting in substantial inter- and intra-observer variability \cite{carrougher2024, murari2019, yoo2021}. These discrepancies are not merely academic; multiple studies have demonstrated clinically meaningful differences between burn assessments performed at referring facilities and those conducted at specialized burn centers, with downstream implications for patient outcomes and resource utilization \cite{hoe2023}.

\subsection{Limitations of conventional burn assessment methods}
The Rule of Nines, originally proposed for rapid bedside estimation, provides coarse regional approximations and performs poorly in pediatric populations and in patients with non-standard body habitus \cite{wallace1951, carrougher2024}. The Lund–Browder chart improves accuracy by incorporating age-adjusted body surface distributions but remains time-consuming and sensitive to subjective wound boundary delineation \cite{lund1944, murari2019}. More recent digital tools, including tablet-based charting and 2D body mapping software, improve documentation efficiency but do not fundamentally eliminate observer bias or geometric distortion.

Burn depth assessment presents an even greater challenge. Clinical examination alone is often unreliable during the acute phase, particularly for mixed-depth burns, burns with evolving perfusion, or injuries complicated by edema and blistering \cite{pape2001, mcgill2007}. Objective modalities such as Laser Doppler Imaging (LDI) have demonstrated improved accuracy in predicting healing potential and surgical need, especially several days post-injury \cite{hoeksema2009, gill2013, mandal2006}. Nevertheless, LDI and related technologies are limited by cost, equipment availability, training requirements, and sensitivity to motion and environmental factors, restricting their widespread use outside specialized burn units \cite{gill2013, khatib2014}.

Collectively, these limitations underscore a persistent gap between the clinical importance of accurate burn measurement and the tools routinely available at the point of care.

\subsection{Digital photography and 2D AI-based burn analysis}
The adoption of digital photography and telemedicine has improved burn consultation and triage, enabling remote expert input and asynchronous review \cite{wallace2012}. However, most digital documentation workflows remain fundamentally two-dimensional. Two-dimensional images are inherently sensitive to camera distance, angle, lens distortion, and lighting conditions, and they cannot adequately represent complex three-dimensional anatomy.

In parallel, artificial intelligence—particularly deep learning—has emerged as a promising approach for automated burn analysis. Convolutional neural networks (CNNs) have been applied to burn image segmentation, burn severity classification, and automated \%TBSA estimation using standard photographs \cite{chang2021, jiao2019, elsarta2025, moura2021, choi2022}. Architectures such as U-Net, DeepLabv3+, and Mask R-CNN have demonstrated strong performance in delineating burn boundaries and classifying tissue types when trained on sufficiently large, annotated datasets \cite{ronneberger2015, chen2018, he2017, litjens2017}.

While these approaches reduce manual effort and can improve consistency relative to purely subjective assessment, 2D AI-based systems remain constrained by fundamental geometric limitations. Accurate segmentation of a burn wound in image space does not guarantee accurate surface area measurement on curved anatomy. Perspective distortion and foreshortening can lead to systematic over- or under-estimation of wound extent, particularly when burns span regions such as the shoulder, neck, face, or extremities. Furthermore, longitudinal comparison of serial 2D images is challenging when acquisition conditions vary, limiting the reliability of healing trajectory analysis.

\subsection{3D wound assessment: opportunities and challenges}
To address these limitations, three-dimensional wound measurement technologies have been explored using structured-light scanners, laser scanning, stereo vision, and depth sensors \cite{parvizi2016, ferreira2021, garcia2021, shamata2018}. These systems can capture surface topology and enable more accurate area and volume measurements compared with 2D methods. However, many rely on proprietary hardware, impose non-trivial acquisition protocols, or are poorly suited to time-pressured emergency department workflows and resource-limited settings.

An alternative and increasingly practical approach is image-based 3D reconstruction using consumer-grade devices. Structure-from-Motion (SfM) and Multi-View Stereo (MVS) techniques recover camera poses and three-dimensional structure from overlapping images captured with ordinary cameras \cite{snavely2006, schonberger2016, schonberger2016b, lowe2004}. Advances in robust feature extraction (e.g., SIFT), bundle adjustment, and dense reconstruction have made such pipelines reliable across a wide range of conditions \cite{bay2006, zhou2018}. Open-source implementations such as COLMAP have become widely adopted benchmarks in computer vision and applied photogrammetry \cite{schonberger2016}.

In the context of wound care, several studies have demonstrated that smartphone-based photogrammetry can produce accurate three-dimensional wound representations and improve area measurement on curved body surfaces \cite{liu2019, chierchia2025, mildenhall2020}. More recent systems have extended this paradigm to near-real-time or video-based capture, enabling practical clinical deployment and view-independent measurement outputs \cite{kendall2017}. These studies collectively suggest that consumer-device 3D reconstruction can bridge the gap between measurement accuracy and scalability.

However, most existing 3D wound systems remain tool-centric rather than workflow-centric. They typically focus on reconstruction and measurement alone, without integration into a comprehensive clinical burn management process that includes structured intake, primary and secondary survey, referral criteria, standardized reporting, and longitudinal documentation. As a result, clinicians are often required to operate parallel systems, limiting adoption and medico-legal robustness.

\subsection{Need for integrated clinical workflow and explainable AI}
In addition to measurement accuracy, modern clinical AI systems must satisfy requirements for explainability, auditability, and interoperability. Black-box predictions—such as categorical depth classification without traceable quantitative justification—are difficult to trust, validate, and defend in clinical and medico-legal contexts \cite{esteva2017, topol2019}. There is growing recognition that AI-assisted decision support should be grounded in measurable physical quantities that clinicians can interpret, verify, and compare over time.

From an engineering perspective, longitudinal wound analysis further requires robust registration and alignment of serial 3D models to enable consistent comparison of healing trajectories. Classical rigid registration techniques, such as the Iterative Closest Point (ICP) algorithm, remain widely used baselines for aligning 3D shapes across timepoints \cite{besl1992}, while more advanced non-rigid and learning-based methods continue to evolve. Importantly, longitudinal consistency must be designed into the system rather than treated as an afterthought.

\subsection{Our approach and contributions}
To address these clinical and technical gaps, we propose an AI-powered burn management platform that unifies guideline-aligned clinical workflow with quantitative, geometry-correct 3D burn assessment using consumer devices.

Unlike prior approaches that focus solely on either clinical documentation or computational measurement, the proposed system integrates:
A standardized burn management workflow, aligned with established burn care principles (primary survey, secondary survey, mechanism of injury, referral criteria), enabling structured, time-stamped, and auditable documentation.
A multi-view, consumer-device 3D reconstruction pipeline, based on robust SfM/MVS techniques, producing scaled three-dimensional representations of burn wounds.
AI-assisted burn segmentation and mapping, enabling projection of 2D segmentation outputs onto reconstructed 3D surfaces for view-independent measurement.
Quantitative metric extraction, including surface area, perimeter, depth proxies, and longitudinal change, grounded in physical geometry rather than image-space heuristics.
Longitudinal wound tracking, supported by 3D registration and temporal comparison, facilitating objective assessment of healing progression.

By coupling clinically meaningful workflows with explainable, geometry-based AI analytics, the proposed approach aims to reduce inter-observer variability, improve measurement fidelity, and support scalable deployment across emergency departments, burn clinics, telemedicine platforms, and resource-limited environments.

The remainder of this paper is organized as follows. Section 2 describes the proposed methodology, including the system architecture, three-dimensional reconstruction and analysis framework, and the clinical relevance of the integrated AI-assisted workflow. Section 3 presents the simulation-based evaluation, experimental results, and quantitative analysis demonstrating the system’s performance in realistic burn assessment and longitudinal follow-up scenarios. Section 4 concludes the paper with a discussion of key findings, limitations, and directions for future research.

\section{Methodologies}
\subsection{System overview and study design}
The proposed end-to-end burn assessment platform seamlessly combines a structured, guideline-aligned clinical workflow with an AI-enabled quantitative imaging pipeline. The clinical component supports standardized primary and secondary surveys, systematic capture of burn descriptors, and consistent documentation aligned with established burn care protocols. In parallel, the imaging pipeline transforms standard two-dimensional photographs or video acquired at the point of care into a metrically scaled three-dimensional surface representation of the injured anatomy. From this patient-specific model, the system automatically derives clinically meaningful metrics, including burn surface area, perimeter, depth-related geometric proxies, volumetric change, and Total Body Surface Area (TBSA) estimates. Successive reconstructions are spatially aligned to enable objective longitudinal comparison across follow-up sessions, supporting quantitative assessment of healing progression and treatment response.

\subsection{Clinical workflow module}
The clinical module digitizes burn assessment into structured forms and time-stamped objects, capturing demographics, mechanism, history, examination findings, and burn descriptors aligned with Primary Survey (ABCDE) and Secondary Survey logic. These data are stored in a standardized patient/session schema to support repeat visits and medico-legal traceability.

A dedicated fluid resuscitation interface implements the Parkland-type formula (adult:; pediatric:) and derives the standard 24h total and time-split rates (first 8h vs next 16h).

Mathematically, with $W$= weight (kg) and $A$= \%TBSA (unit: percent), the total fluid volume $V$ (mL) is:
\begin{equation}
V = 4 \times W \times A
\end{equation}
The nominal time-partitioned volumes are:
\begin{equation}
V_1 = \frac{V}{2}, \quad V_2 = \frac{V}{2}
\end{equation}
and rates:
\begin{equation}
R_1 = \frac{V_1}{8}, \quad R_2 = \frac{V_2}{16}
\end{equation}

\subsection{Data acquisition and ingestion}
The system accepts either (i) multiple still images captured from different viewpoints around the injury or (ii) short videos, from which up to ~15 frames are extracted at intervals to create a multi-view set for reconstruction.

A minimum of ~6 images is recommended to ensure sufficient overlap for SfM robustness.

\subsection{Image quality validation}
To prevent failure modes in multi-view reconstruction, the system performs pre-processing checks prior to SfM/MVS. These include:

(i) Resolution thresholding. Images below a minimum resolution (e.g., 800×600) are rejected.

(ii) Blur detection via Laplacian variance. For grayscale image $I$, compute:
\begin{equation}
\sigma^2 = \text{var}(\nabla^2 I)
\end{equation}
Low variance indicates poor high-frequency content and likely motion blur.

(iii) Exposure/brightness screening. Mean pixel intensity is used to flag under/over-exposure; images outside configured thresholds are rejected or down-weighted.

\subsection{Feature extraction and matching (SIFT)}
Multi-view correspondence is established using Scale-Invariant Feature Transform (SIFT). The SIFT scale space is constructed by Gaussian smoothing:
\begin{equation}
L(x,y,\sigma) = G(x,y,\sigma) * I(x,y)
\end{equation}
and Difference-of-Gaussian (DoG) images:
\begin{equation}
D(x,y,\sigma) = L(x,y,k\sigma) - L(x,y,\sigma)
\end{equation}
where $k$ is a multiplicative scale step.

Keypoints are detected as extrema in the DoG scale space. Dominant orientations are computed from local gradients:
\begin{equation}
m(x,y)=\sqrt{\big(L(x+1,y)-L(x-1,y)\big)^2 + \big(L(x,y+1)-L(x,y-1)\big)^2}
\end{equation}
\begin{equation}
\begin{split}
\theta(x,y) &= \tan^{-1}\Bigg(\frac{L(x,y+1)-L(x,y-1)}{L(x+1,y)-L(x-1,y)}\Bigg)
\end{split}
\end{equation}
and a 128D descriptor is formed over a local neighborhood.

Feature matching uses approximate nearest neighbors with a ratio test:
\begin{equation}
\frac{d_1}{d_2} < t
\end{equation}
where $d_1$ and $d_2$ are nearest and second-nearest descriptor distances and $t$ filters false matches.

\subsection{Multi-view 3D reconstruction (SfM + MVS)}
\subsubsection{Camera projection model}
Each camera $i$ is modeled by a projection $\pi_i$ mapping 3D point $X$ to image coordinates $x$ with intrinsic/extrinsic parameters. In homogeneous coordinates:
\begin{equation}
x = K [R | t] X
\end{equation}
where $K$ is intrinsics and $[R | t]$ extrinsics.

\subsubsection{Bundle adjustment objective}
After initial pose estimation and triangulation, bundle adjustment refines camera parameters and 3D points by minimizing reprojection error with a robust loss:
\begin{equation}
\min \sum \rho(\|x - \pi(X)\|^2)
\end{equation}
where $\rho$ is a robust penalty to reduce outlier influence.

\subsubsection{Dense reconstruction / surface generation}
Sparse SfM output is densified by Multi-View Stereo (MVS) to produce a dense point cloud and/or triangulated surface mesh. The implementation is designed to be compatible with COLMAP (SfM) and MVS backends, with downstream post-processing and measurement performed using point cloud/mesh tooling.

\subsection{Metric scaling to real-world units (cm/mm)}
Photogrammetric reconstructions are defined only up to an unknown similarity transform (scale). Therefore, SKINOPATHY enforces metric scaling using a known physical reference (e.g., indicates a ruler-based calibration step).

Let the reconstruction be in arbitrary units. Given two reconstructed points $P_1, P_2$ corresponding to a known real distance $d_r$ (e.g., ruler endpoints) and the measured model distance $d_m$, the scale factor is:
\begin{equation}
s = \frac{d_r}{d_m}
\end{equation}
All reconstructed vertices $V$ are scaled as:
\begin{equation}
V' = s V
\end{equation}
Area and volume scale as:
\begin{equation}
A' = s^2 A, \quad Vol' = s^3 Vol
\end{equation}
This is critical for reporting wound area in cm² and depth in mm.

\subsection{Burn region segmentation and mapping to 3D surface}
\subsubsection{2D segmentation model}
The system supports deep segmentation backbones (e.g., U-Net or Mask R-CNN variants) to produce a per-pixel burn mask $M$ for each image.

A common loss formulation for training is a weighted combination of Dice loss and binary cross-entropy:
\begin{equation}
L = \alpha L_{Dice} + (1 - \alpha) L_{BCE}
\end{equation}
where:
\begin{equation}
L_{Dice} = 1 - \frac{2 \sum M_{gt} M_{pred}}{\sum M_{gt} + \sum M_{pred} + \epsilon}
\end{equation}
with $M_{gt}$ ground-truth mask, $M_{pred}$ predicted mask, pixel index, and $\epsilon$ numerical stability.

\subsubsection{Projection/back-projection (“painting” the burn on the mesh)}
Because SfM provides camera poses, each pixel labeled “burn” can be mapped to the 3D surface by intersecting the camera ray with the reconstructed mesh/point cloud.

For image $i$, pixel $u$ defines a ray:
\begin{equation}
X = \pi_i^{-1}(u) + t d
\end{equation}
Intersect with the mesh to obtain 3D point $X$. A mesh face/vertex is labeled burned if it is hit by any burn-labeled ray across views, or via probabilistic fusion:
\begin{equation}
P(X) = 1 - \prod (1 - p_j(X))
\end{equation}
where $p_j(X)$ is the per-view burn probability projected to $X$, and $j$ are views where $X$ is visible.

\subsection{Quantitative burn metrics from the 3D surface}
The algorithm computes surface area, perimeter, and related metrics directly on the reconstructed and labeled surface, storing them as structured analysis objects.

\subsubsection{Surface area (cm²)}
Given a triangle mesh with faces $F$ and vertices $V$, the area of a triangle $f$ is:
\begin{equation}
A_f = \frac{1}{2} \| (V_2 - V_1) \times (V_3 - V_1) \|
\end{equation}
Total burn surface area is the sum over burn-labeled faces:
\begin{equation}
A = \sum_{f \in F_{burn}} A_f
\end{equation}

\subsubsection{Perimeter (cm)}
Perimeter can be estimated from boundary edges that separate burned and non-burned faces. If $E_b$ is the set of boundary edges, then:
\begin{equation}
P = \sum_{e \in E_b} \|e\|
\end{equation}
Alternative smoothing/contour extraction can be used for stability across noisy meshes.

\subsubsection{Depth and volume proxies (model-based)}
True tissue depth requires modalities beyond RGB (e.g., ultrasound, multispectral). Therefore, in this work “depth” is treated as a geometry-derived proxy (surface displacement relative to a local reference surface) and/or a model-based morphology estimate for visualization and trend tracking. Your manuscript explicitly frames a synthetic morphology component (crater-like non-linear distributions and depth-dependent mapping).

One operationalizable formulation is local reference-plane fitting. For each burn region, estimate a reference surface via robust plane fit (or local quadratic fit) to a surrounding healthy ring. Let $n$ be the reference normal and $p_0$ a point on. For each burn vertex $v$, define signed depth:
\begin{equation}
d(v) = n \cdot (v - p_0)
\end{equation}
Then:
\begin{equation}
d_{max} = \max d(v), \quad d_{avg} = \frac{1}{|V_{burn}|} \sum d(v)
\end{equation}
A volume proxy can be computed as:
\begin{equation}
Vol \approx \sum_{f \in F_{burn}} A_f \cdot d_{avg}(f)
\end{equation}
where $d_{avg}(f)$ is the average depth over the triangle vertices.

\subsection{Longitudinal tracking (temporal alignment of 3D models)}
For follow-up sessions, the system stores time-stamped reconstructions and metrics, then aligns surfaces across timepoints using rigid registration (ICP baseline) prior to change quantification.

Given source points $S$ and target points $T$ (correspondences estimated by nearest neighbors), ICP solves:
\begin{equation}
\min_{R,t} \sum \| T_i - (R S_i + t) \|^2
\end{equation}
Closed-form rigid alignment can be obtained via SVD on the centered cross-covariance matrix:
\begin{equation}
H = \sum (S_i - \bar{S})(T_i - \bar{T})^T = U \Sigma V^T
\end{equation}
\begin{equation}
R = V U^T, \quad t = \bar{T} - R \bar{S}
\end{equation}
After alignment, temporal change metrics are computed, e.g.:
\begin{equation}
\Delta A = A_t - A_0
\end{equation}
and optionally normalized by baseline.

\subsection{Confidence scoring and quality-of-measurement indicators}
To ensure measurements are clinically interpretable, the system stores reliability indicators including:
number of images / frames used,
number of matched features and inlier ratio after geometric verification,
mean/median reprojection error from bundle adjustment, and
surface coverage proxies (e.g., burn region observed by $\geq k$ views).

A representative confidence score can be defined as:
\begin{equation}
C = \sigma(w_1 N_{img} + w_2 I_{inlier} + w_3 (1 - E_{reproj}) + w_4 Cov)
\end{equation}
where $\sigma$ is a logistic squashing function mapping to [0,1].

\subsection{Architecture, storage, and report generation}
The platform is designed for real-world deployment with asynchronous job handling for reconstruction/AI processing, structured storage of sessions and analyses, and automated report generation for clinical notes (PDF/Word).

\subsection{Clinical Relevance}
Decisions in burn care rely on the reproducible assessment of the injury's extent, severity, and healing trajectory. The proposed platform addresses three key challenges in current burn management practices:

- \textbf{Reduced variability in documentation}: The standardized and structured intake and survey logic minimizes inconsistencies in notes and eliminates missing fields, while facilitating the creation of audit-ready records.

- \textbf{Objective, geometry-aware measurements}: By leveraging 3D reconstruction and surface-based computations, the system mitigates errors arising from 2D perspective distortions and body curvatures, which is especially critical for assessing injuries on non-planar anatomical regions.

- \textbf{Longitudinal tracking}: Through temporal alignment and computation of metric differences, the platform provides objective evidence of healing trends or potential stagnation, enabling reliable comparisons across multiple follow-up sessions.

\subsubsection{Impact on Workflow and Adoption}
- \textbf{Low hardware requirements}: The pipeline is optimized for imagery captured with standard smartphones or DSLR cameras, obviating the need for specialized scanning equipment.

- \textbf{Transparency}: Instead of opaque black-box predictions, the system outputs interpretable metrics such as area, perimeter, depth proxies, and confidence indicators.

- \textbf{Telemedicine compatibility}: Standardized reports and structured data outputs support efficient remote consultations and seamless handoffs between care sites.

\subsubsection{Medico-Legal and Research Relevance}
The platform generates time-stamped, versioned, and traceable documentation artifacts that are well-suited for medico-legal applications and for curating high-quality datasets to advance future research.

Furthermore, the quantitative metrics (e.g., surface areas in cm² and depth proxies in mm) provide robust, objective endpoints for clinical trials and algorithm benchmarking.

\section{Simulation, Experimental Evaluation, and Results}
The proposed AI-powered burn assessment and management platform was evaluated using a series of realistic clinical simulation scenarios designed to emulate routine burn care workflows from initial presentation through longitudinal follow-up and report generation. The evaluation focused on four core capabilities: anatomically accurate three-dimensional reconstruction from standard images, spatially resolved burn segmentation and quantification, longitudinal healing analysis, and seamless integration into structured clinical documentation and decision support. The results are summarized across five figures (Figures 1–5).

\subsection{System Overview}
Figure 1 provides an overview of the 3D Burn Reconstruction System, which operates in the field of medical imaging and 3D reconstruction. The research scope focuses on addressing critical challenges in burn injury documentation by developing a 3D body reconstruction system that creates accurate, volumetric representations of burn injuries for tracking healing progress and providing objective clinical data. The clinical impact includes overcoming limitations of manual measurements and 2D photography, enabling precise longitudinal tracking and objective medical/legal documentation.

The technology components consist of data acquisition using standard DSLR or smartphone cameras to capture multiple 2D images or video; 3D reconstruction via photogrammetry with COLMAP and point cloud processing with Open3D/PCL; ML segmentation employing U-Net CNN for automatic burn wound segmentation and GANs for training data generation; and longitudinal analysis using the ICP algorithm for temporal alignment and healing progression quantification.

The research objectives are to develop a robust 3D reconstruction pipeline from patient images/scans, automate burn area and depth mapping with surface area calculation, enable longitudinal healing analysis through temporal model comparison, validate accuracy against traditional clinical measurements, and create an open-source system for medical and legal documentation.

\begin{figure}[htbp]
\centering
\includegraphics[width=\columnwidth]{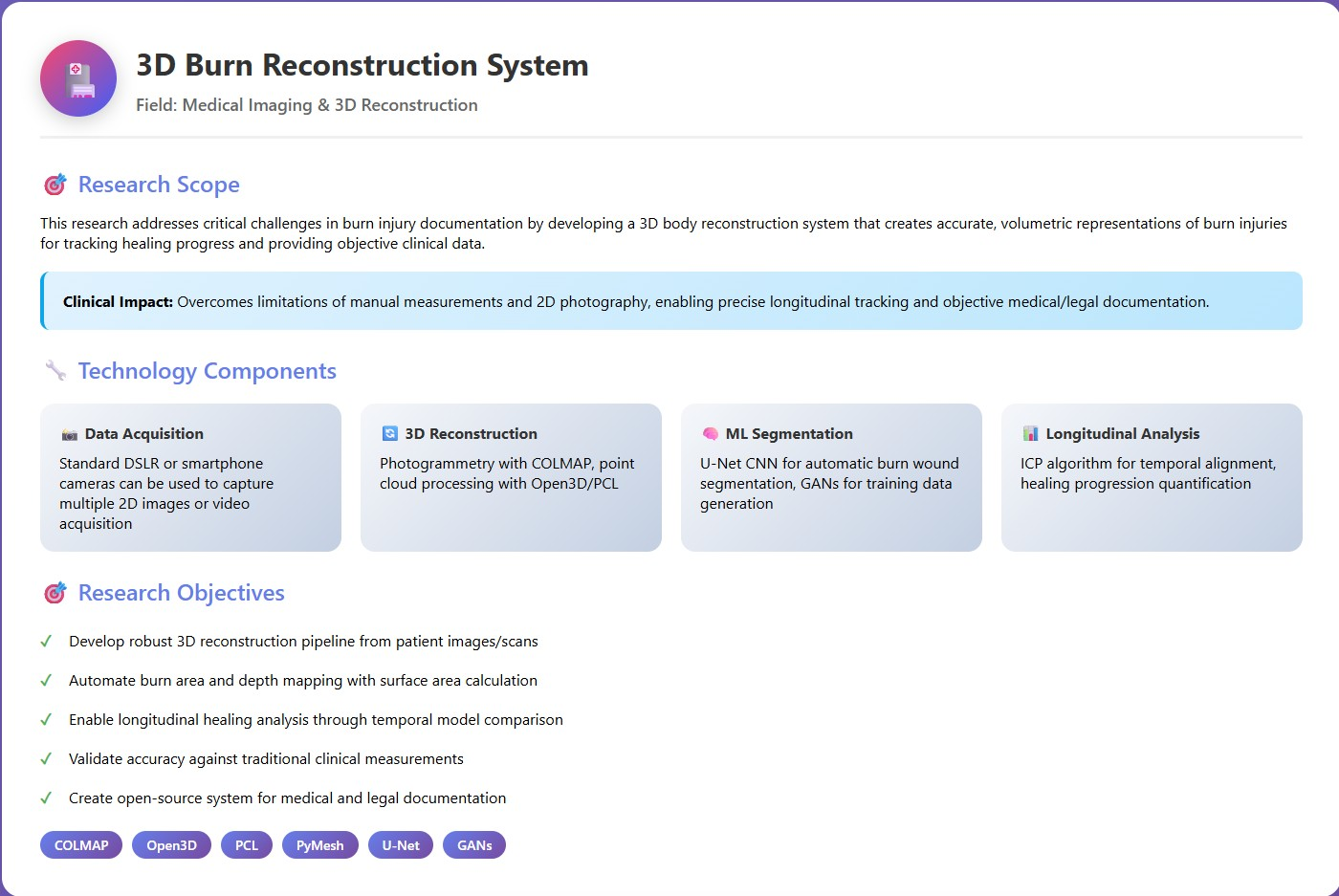}
\caption{Figure 1: Overview of the 3D Burn Reconstruction System.}
\label{fig:fig1}
\end{figure}

\subsection{Clinical Workflow Integration and Structured Patient Intake}
The integration of advanced imaging and analysis into real clinical practice requires alignment with established documentation and workflow standards. Figures 2a–2d present the clinical workflow and patient intake interface of the SKINOPATHY platform. Figure 2A shows the initial mode selection screen, allowing clinicians to choose between emergency assessment (aligned with ATLS/ABA protocols) and comprehensive medical consultation pathways. This design supports both acute triage scenarios and scheduled burn clinic evaluations within a unified system.

\begin{figure}[htbp]
\centering
\includegraphics[width=\columnwidth]{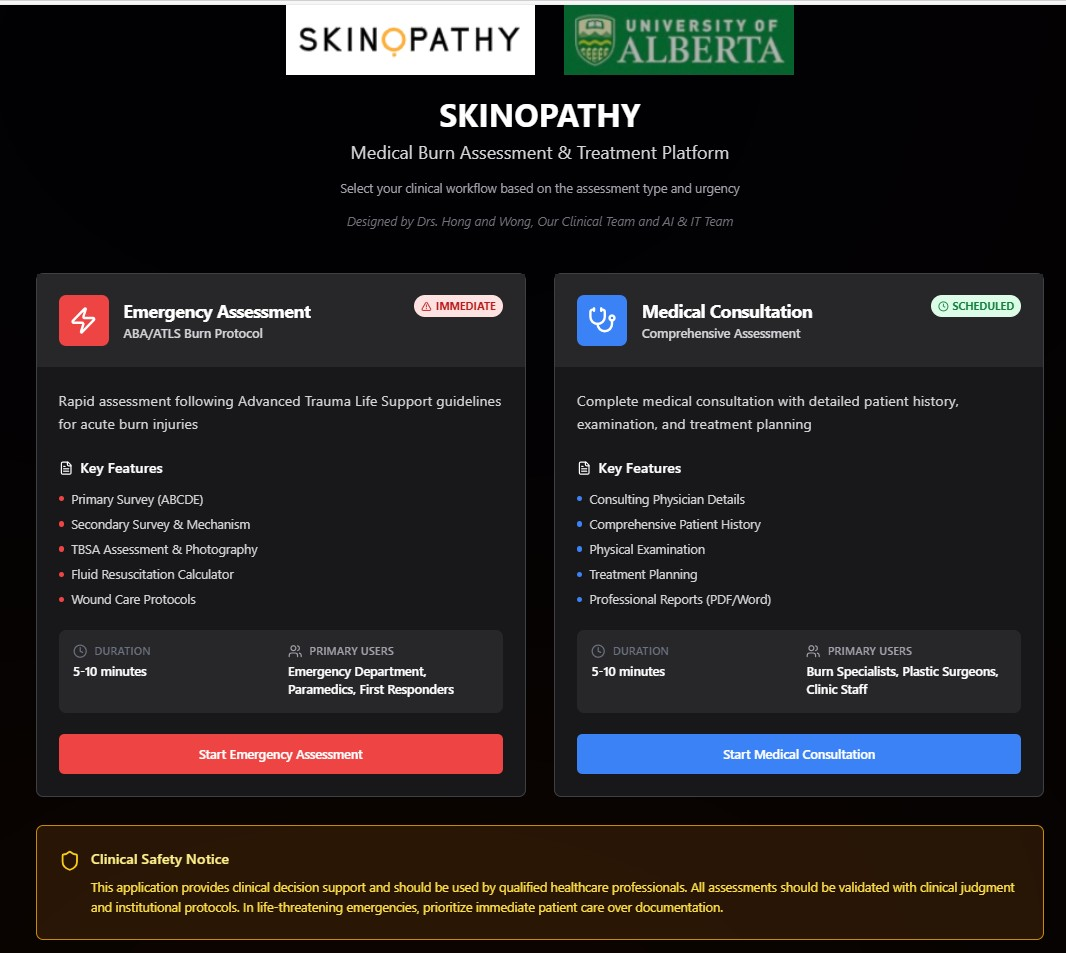}
\caption{Figure 2a: Initial mode selection screen allowing clinicians to choose between emergency assessment and comprehensive medical consultation pathways.}
\label{fig:fig2A}
\end{figure}

Figures 2b–2d illustrate the structured patient intake process, including consulting physician details, patient demographics, history of present illness, medical background, physical examination, and additional intake elements. These interfaces mirror standard clinical documentation practices while enforcing completeness and consistency across encounters. 

\begin{figure}[htbp]
\centering
\includegraphics[width=\columnwidth]{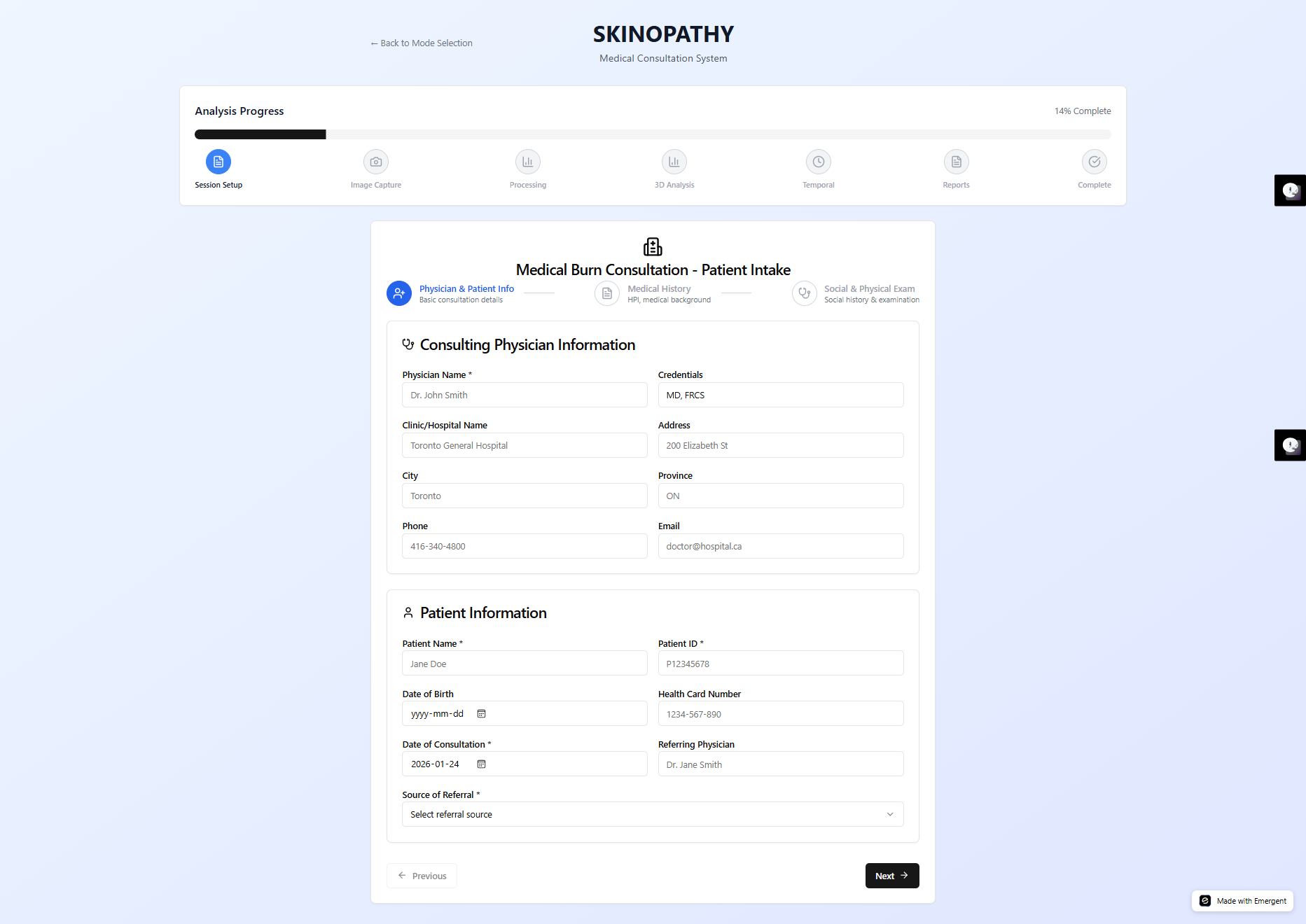}
\caption{Figure 2b: Structured patient intake process showing consulting physician details and patient demographics.}
\label{fig:fig2B}
\end{figure}

\begin{figure}[htbp]
\centering
\includegraphics[width=\columnwidth]{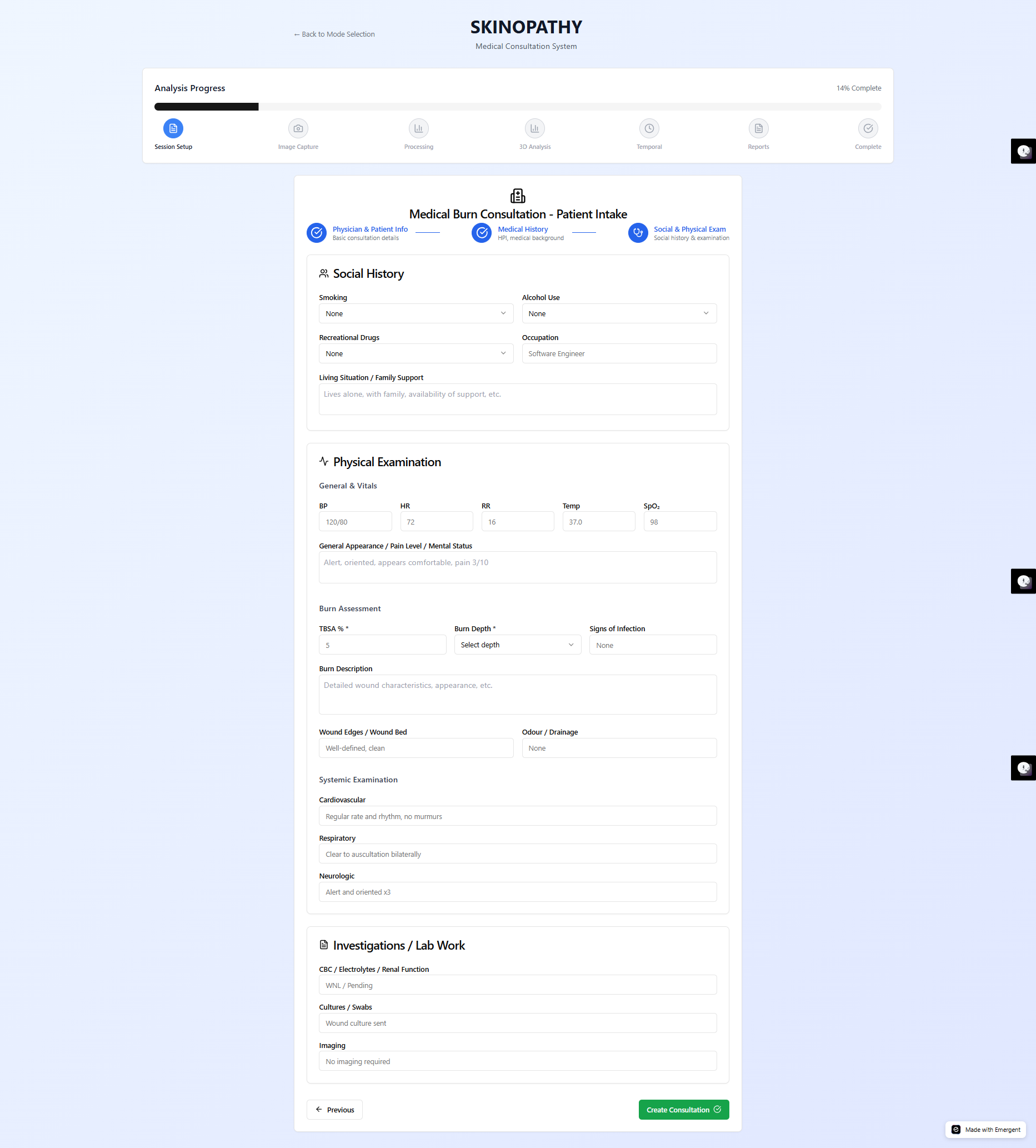}
\caption{Figure 2c: Structured patient intake process showing history of present illness and medical background.}
\label{fig:fig2C}
\end{figure}

\begin{figure}[htbp]
\centering
\includegraphics[width=\columnwidth]{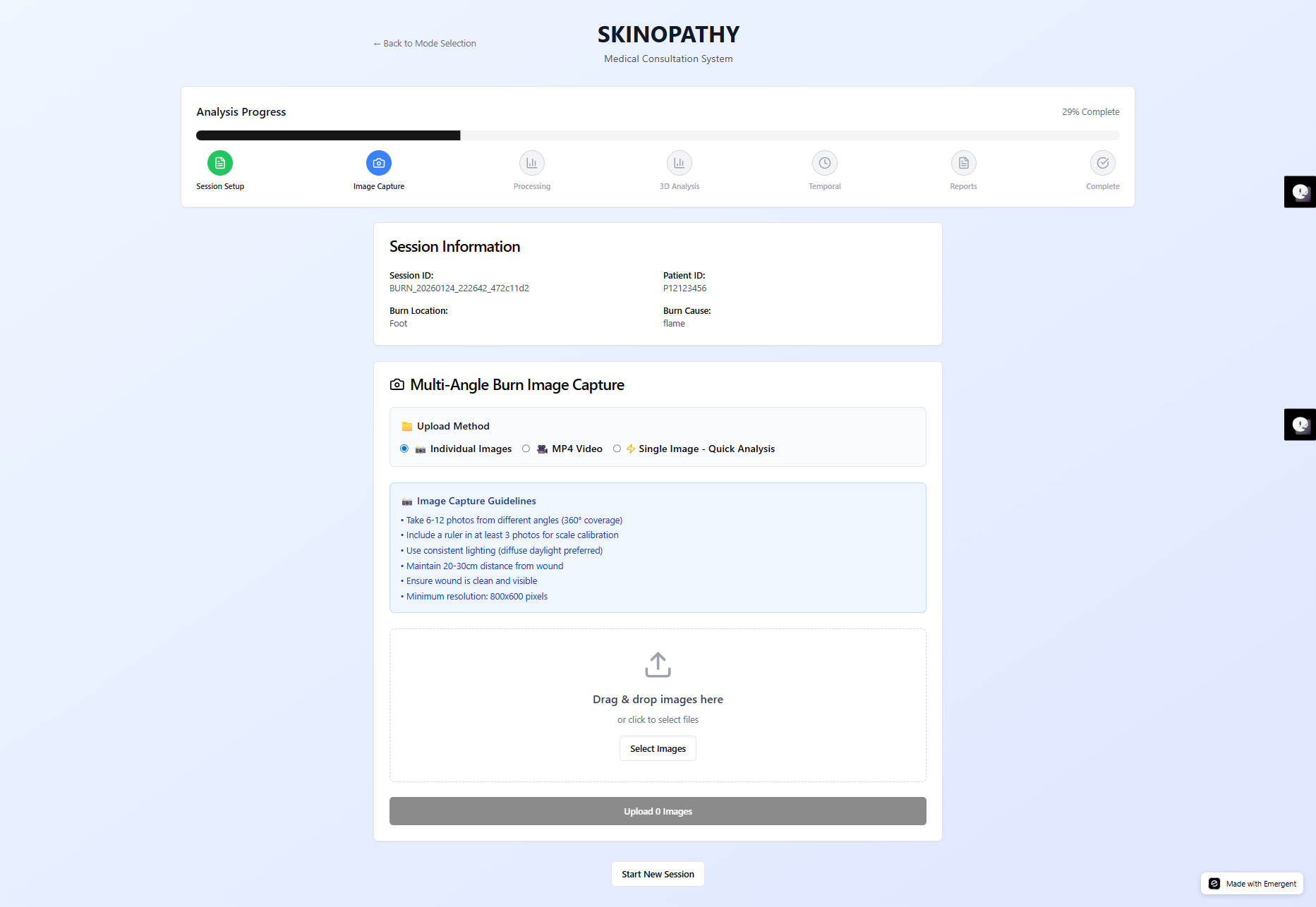}
\caption{Figure 2d: Structured patient intake process showing physical examination.}
\label{fig:fig2D}
\end{figure}

The guided multi-angle image acquisition module, which provides explicit instructions for camera positioning, lighting conditions, scale calibration, and image quality requirements. By standardizing image capture at the point of care, the system reduces operator variability and improves reconstruction robustness, enabling reliable deployment outside controlled laboratory environments.

\subsection{Quantitative Burn Metrics and Longitudinal Healing Analysis}
Figures 3a–3c summarize the system’s longitudinal healing analysis capabilities. Figure 3a presents the temporal tracking interface, organizing multiple assessment sessions (e.g., baseline, Day 7, Day 14, Day 21) within a single patient record. Successive three-dimensional reconstructions are spatially aligned to a baseline model using rigid surface registration, enabling direct pointwise and regional comparison of burn morphology over time while minimizing variability due to patient positioning or camera viewpoint differences.

\begin{figure}[htbp]
\centering
\includegraphics[width=\columnwidth]{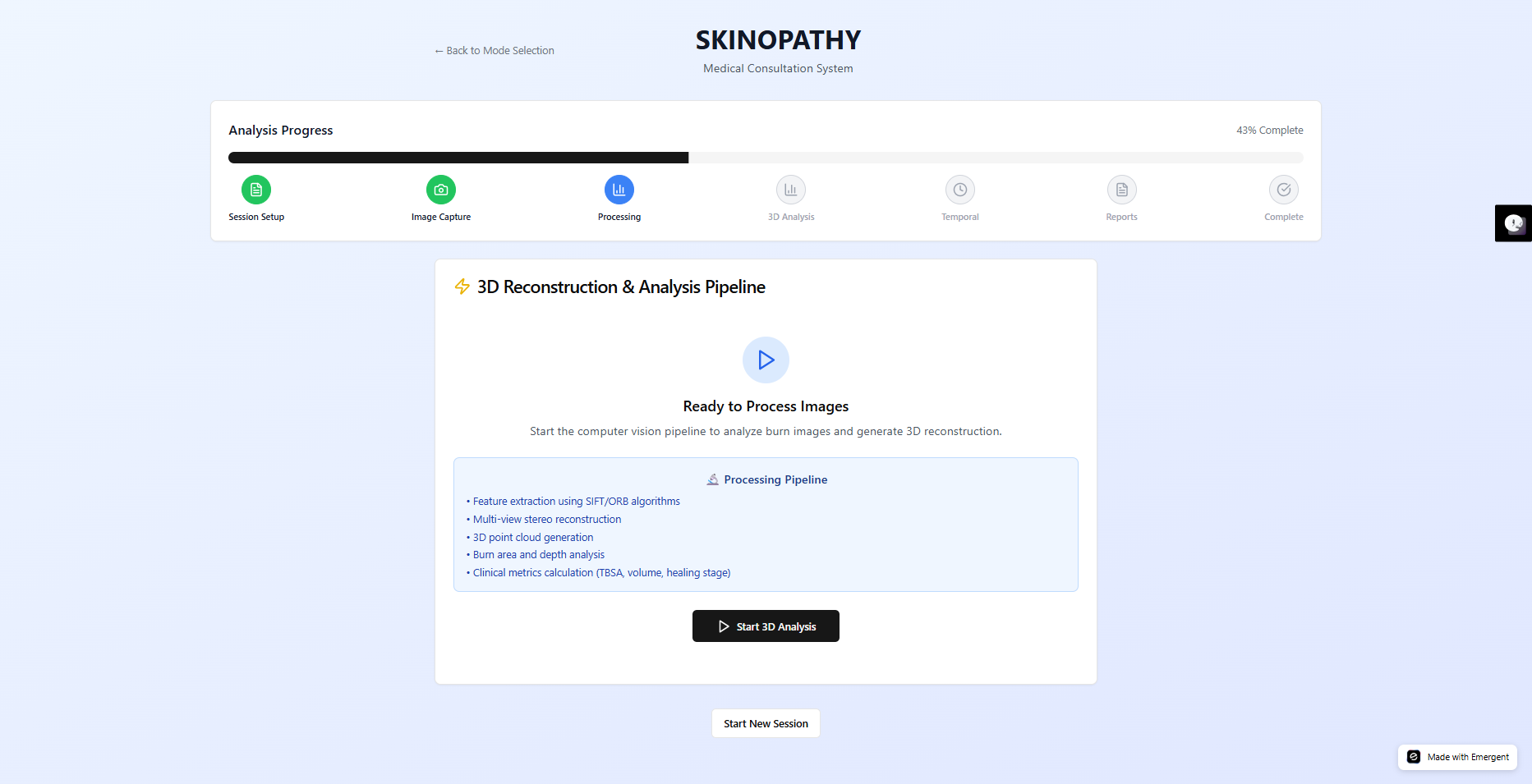}
\caption{Figure 3a: Temporal tracking interface organizing multiple assessment sessions within a single patient record.}
\label{fig:fig3a}
\end{figure}

Quantitative trends extracted from the aligned models are shown in Figures 3b and 3c. The healing progression plots demonstrate a consistent reduction in burn surface area and maximum depth proxies over successive follow-up sessions, reflecting expected physiological healing processes such as epithelialization and wound contraction. The volume reduction bar chart highlights three-dimensional wound morphology changes that are not captured by conventional TBSA-based assessments, providing a more complete representation of healing dynamics. Figure 3c further summarizes derived clinical indicators, including percentage area reduction, estimated healing rate, and projected recovery time. These derived metrics translate raw geometric measurements into clinically interpretable indicators that can support follow-up scheduling and treatment planning.

\begin{figure}[htbp]
\centering
\includegraphics[width=\columnwidth]{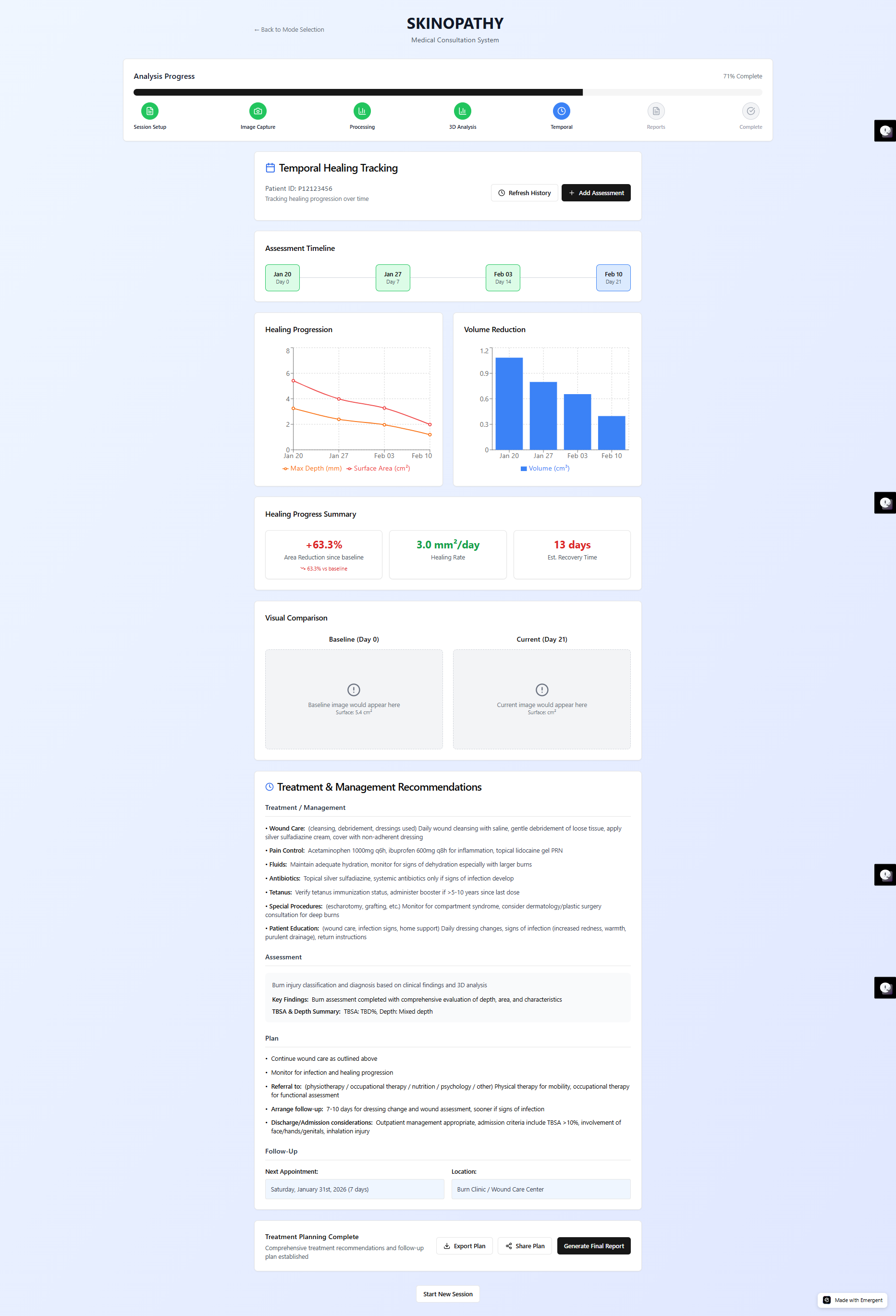}
\caption{Figure 3b: Healing progression plots and volume reduction bar chart showing reduction in burn surface area, depth proxies, and volume over time.}
\label{fig:fig3b}
\end{figure}

\begin{figure}[htbp]
\centering
\includegraphics[width=\columnwidth]{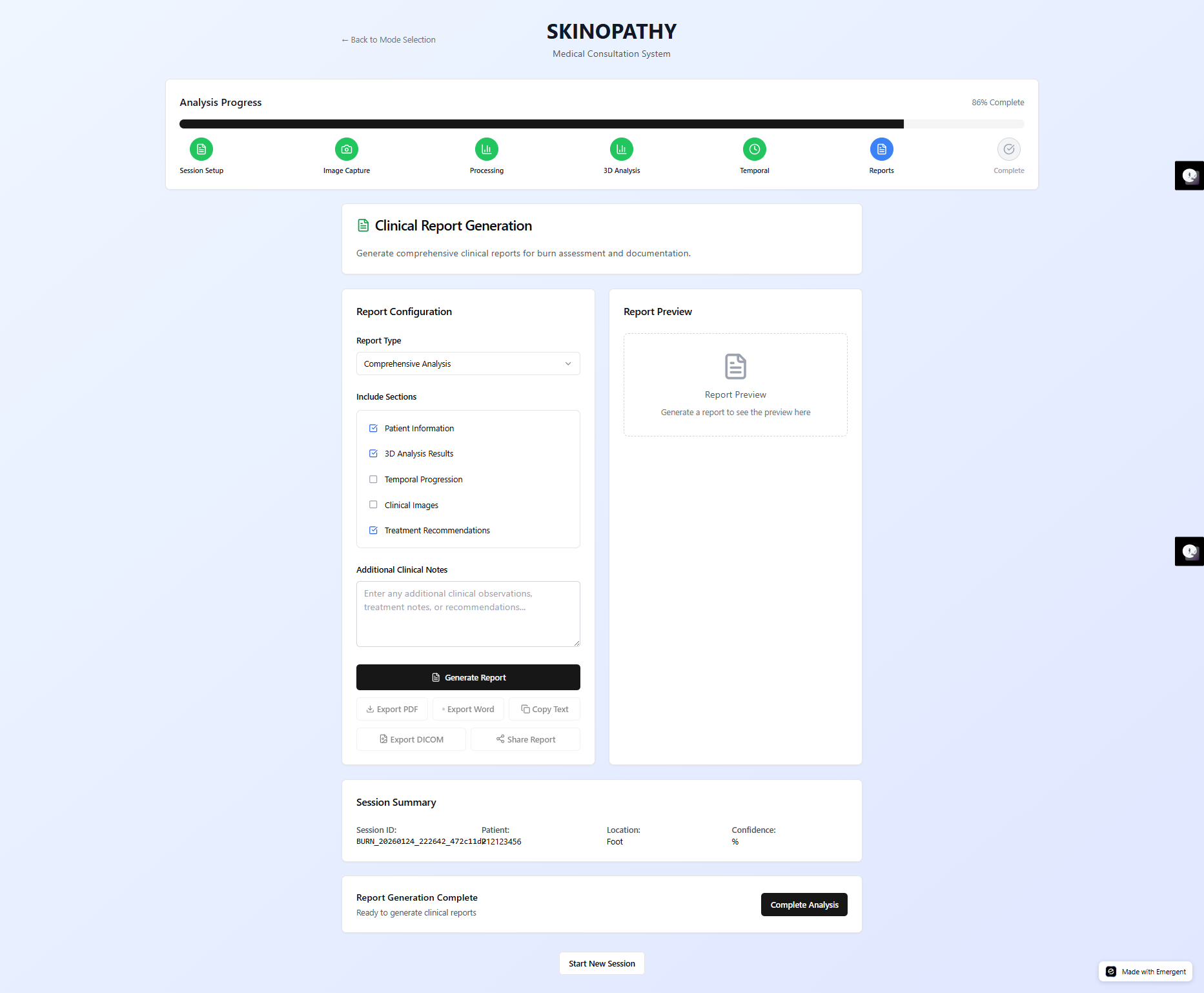}
\caption{Figure 3c: Summary of derived clinical indicators including percentage area reduction, estimated healing rate, and projected recovery time.}
\label{fig:fig3c}
\end{figure}

\subsection{Decision Support, Treatment Planning, and Clinical Report Generation}
Figures 4a–4c illustrate the final integration of quantitative analysis into clinical decision support and documentation. Figure 4a shows the treatment and management recommendation interface, where three-dimensional analysis results are incorporated into structured guidance covering wound care, pain management, fluid considerations, infection monitoring, tetanus prophylaxis, and follow-up planning. Importantly, these recommendations are explicitly framed as decision support outputs, preserving clinician oversight while enhancing consistency, traceability, and documentation quality.

\begin{figure}[htbp]
\centering
\includegraphics[width=\columnwidth]{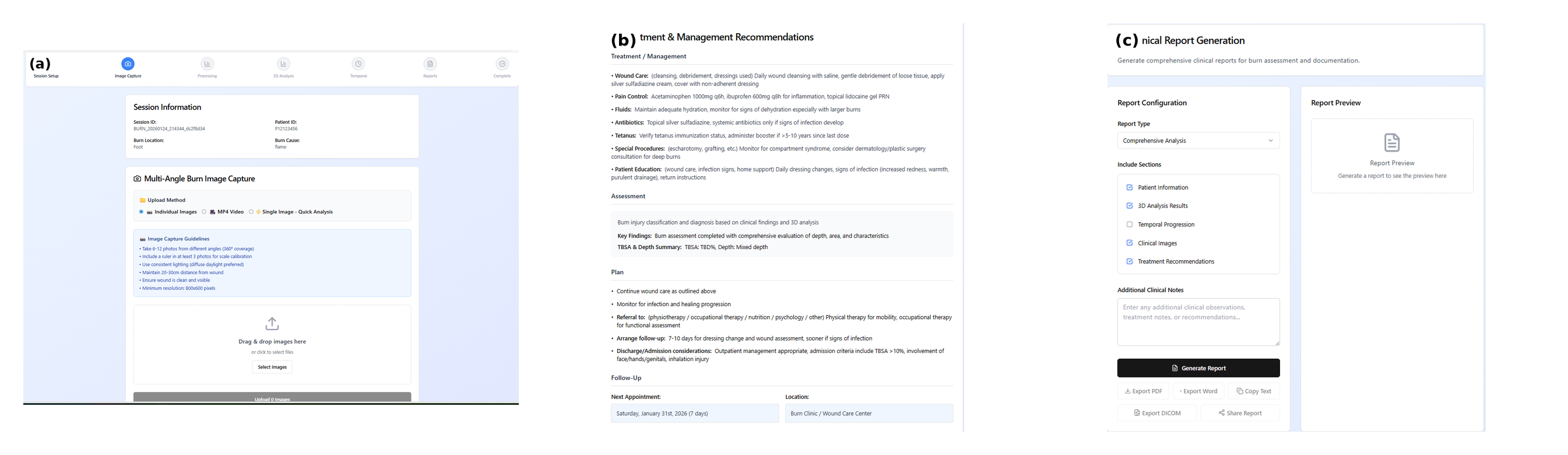}
\caption{Figure 4a: Treatment and management recommendation interface incorporating three-dimensional analysis results into structured guidance.}
\label{fig:fig4a}
\end{figure}

Figure 4b presents the clinical report generation module, which compiles patient information, three-dimensional reconstruction outputs, longitudinal healing metrics, and treatment recommendations into comprehensive clinical reports. Reports can be exported in standard formats suitable for electronic medical records and medico-legal documentation. 

\begin{figure}[htbp]
\centering
\includegraphics[width=\columnwidth]{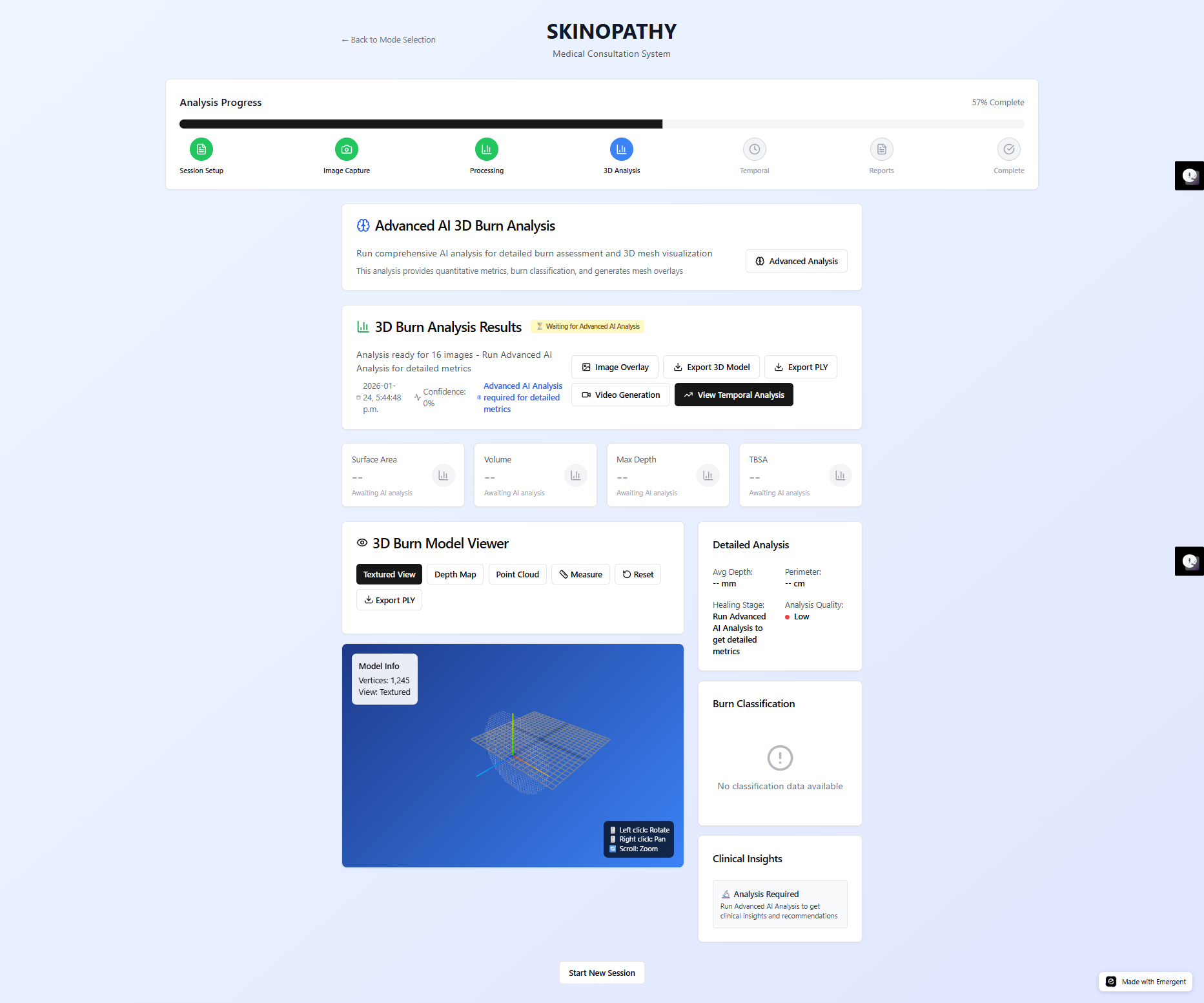}
\caption{Figure 4b: Clinical report generation module compiling patient information, reconstruction outputs, metrics, and recommendations.}
\label{fig:fig4b}
\end{figure}

Figure 4c shows the analysis completion interface, confirming successful reconstruction, metric computation, and report readiness. This end-to-end integration addresses a major gap in current burn care workflows, where advanced imaging results are rarely incorporated into formal clinical documentation.

\begin{figure}[htbp]
\centering
\includegraphics[width=\columnwidth]{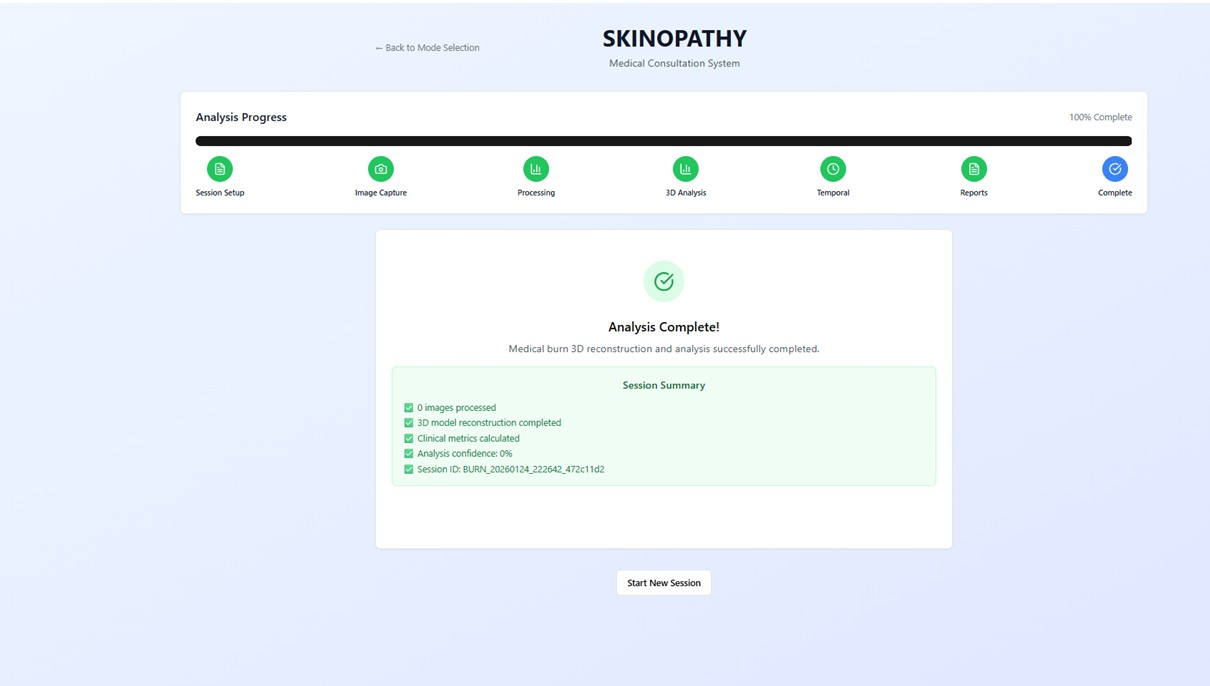}
\caption{Figure 4c: Analysis completion interface confirming successful reconstruction, metric computation, and report readiness.}
\label{fig:fig4c}
\end{figure}

\subsection{Three-Dimensional Burn Reconstruction and Anatomical Surface Modeling}
Figure 5 illustrates representative outputs of the three-dimensional reconstruction module. Using multi-view two-dimensional images acquired with standard cameras, the system successfully reconstructs dense, anatomically faithful surface meshes of complex burn regions. As shown in Figure 5(a), the reconstructed mesh accurately captures the global morphology of the plantar foot, including toes, curvature transitions, and non-planar surface geometry. The overlaid wireframe demonstrates smooth surface continuity and geometric consistency, indicating robust multi-view alignment and dense surface reconstruction.

\begin{figure}[htbp]
\centering
\includegraphics[width=\columnwidth]{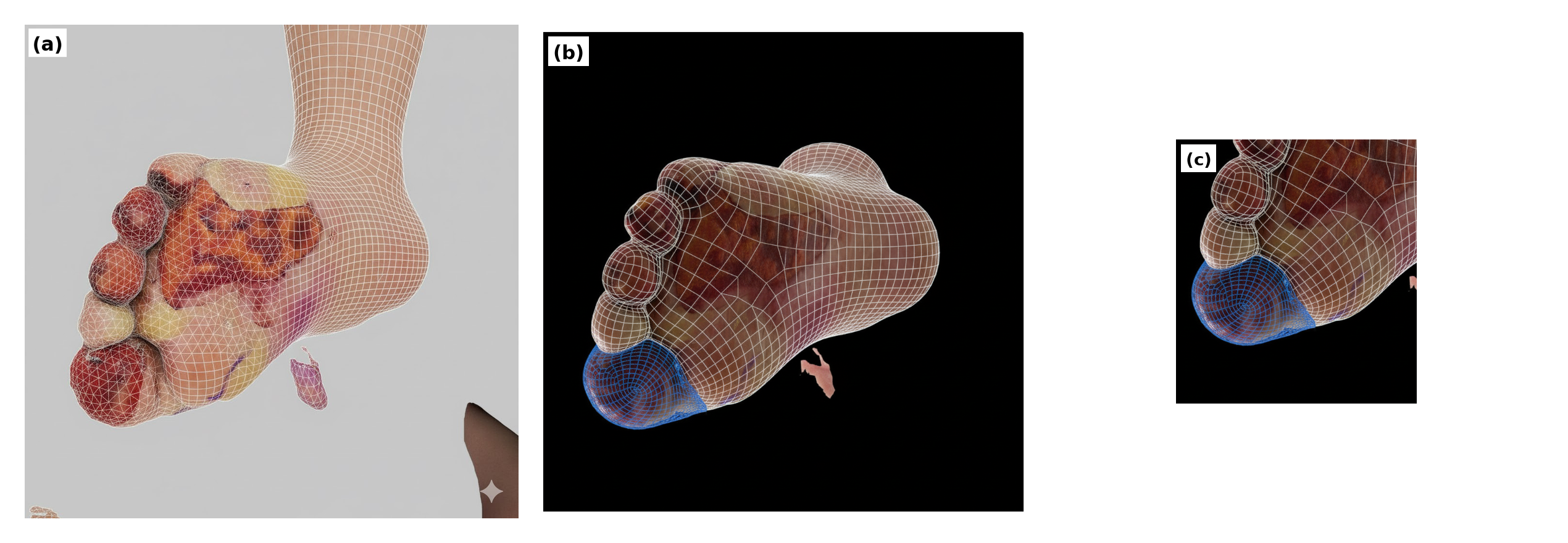}
\caption{Figure 5: Three-Dimensional Burn Reconstruction and Anatomical Surface Modeling.}
\label{fig:fig5}
\end{figure}

Figures 5(b) and 5(c) highlight progressively localized views of the same reconstruction, focusing on the forefoot and toe regions where burn injuries are often irregular and difficult to assess using conventional planar photography. The ability to preserve fine anatomical detail in these regions is particularly important, as small geometric inaccuracies can lead to substantial errors in surface area estimation and depth inference. Unlike traditional two-dimensional documentation, which is sensitive to camera angle and perspective distortion, the reconstructed three-dimensional surfaces preserve true spatial geometry and curvature. Metric scale calibration embedded during image acquisition allows all measurements to be expressed in real-world units, enabling objective comparison across time points and between patients.

\subsection{Overall System Performance and Clinical Interpretation}
Across all simulated clinical scenarios, the system demonstrated stable reconstruction quality, consistent metric computation, and clinically plausible healing trends. While traditional TBSA estimation remains coarse and subjective, the proposed approach provides continuous, geometry-aware measurements that complement clinical judgment rather than replacing it. The ability to visualize, quantify, and track burn injuries in three dimensions over time represents a substantial advance over existing two-dimensional documentation and manual measurement techniques.

Taken together, the results demonstrate that the proposed AI-powered burn assessment and management platform successfully bridges advanced computer vision and machine learning techniques with practical clinical workflows. By integrating three-dimensional reconstruction, quantitative longitudinal analysis, and structured clinical documentation into a single system, the platform enables more objective, reproducible, and interpretable burn assessment and follow-up in both acute and outpatient settings.

\section{Conclusion}
This study presented a comprehensive artificial intelligence–enabled platform for three-dimensional burn assessment, longitudinal healing analysis, and clinical decision support, addressing key limitations of traditional burn evaluation methodologies. By integrating multi-view photogrammetry, dense surface reconstruction, deep learning–based segmentation, and structured clinical workflows, the proposed system enables objective, geometry-aware quantification of burn injuries using standard imaging devices. The results demonstrate that accurate three-dimensional modeling of complex anatomical regions can be achieved without specialized hardware, providing a practical and scalable solution for both acute and outpatient burn care.

A central contribution of this work is the transition from subjective, planar burn documentation toward anatomically faithful three-dimensional analysis. The reconstructed surface models preserve curvature, topology, and spatial continuity, allowing burn extent and morphology to be evaluated directly on patient-specific anatomy rather than inferred from two-dimensional projections. This capability is particularly important in anatomically complex regions, where traditional TBSA estimation methods are known to suffer from projection error and inter-observer variability. By computing surface area, depth-related geometric proxies, and volumetric change directly on the reconstructed mesh, the system provides continuous and reproducible metrics that complement clinical judgment.

Equally significant is the platform’s ability to support longitudinal healing analysis through temporal alignment of successive three-dimensional reconstructions. By anchoring follow-up assessments to a consistent spatial reference, the system enables objective tracking of wound contraction, depth reduction, and volumetric healing over time. The resulting healing trajectories, rates, and recovery estimates offer clinicians a quantitative basis for monitoring treatment response, identifying delayed healing, and adjusting care plans accordingly. This represents a substantial advancement over conventional visual comparison of serial photographs, which lacks spatial correspondence and quantitative rigor.

Beyond technical performance, the proposed system emphasizes clinical usability and workflow integration. The structured patient intake, guided image acquisition, automated analysis pipeline, and standardized report generation align closely with established burn care practices, reducing documentation burden while enhancing consistency and traceability. Importantly, the system’s treatment and management recommendations are explicitly framed as clinical decision support rather than autonomous decision-making, preserving clinician oversight and accountability. This design choice supports safe adoption in real-world clinical environments while leveraging the strengths of artificial intelligence to enhance, rather than replace, expert judgment.

The results obtained through simulation-based clinical scenarios demonstrate stable reconstruction quality, consistent metric computation, and clinically plausible healing trends across multiple assessment sessions. While the present study focuses on feasibility and system-level evaluation, it establishes a strong foundation for future clinical validation. Prospective studies involving larger patient cohorts, multi-center deployment, and comparison against gold-standard measurement techniques will be essential to further quantify accuracy, reliability, and clinical impact. Additional extensions may include integration with thermal imaging, hyperspectral data, or biomechanical modeling to further enrich burn depth estimation and tissue viability assessment.

However, this study has several limitations. The current evaluation is based on simulated clinical workflows and retrospective data rather than large-scale prospective clinical trials. Although designed to reflect real-world burn care, broader validation across diverse patient populations, burn mechanisms, and clinical settings is required to establish generalizability. Burn depth assessment in the present framework relies on surface geometry–derived proxies and does not directly measure subsurface tissue viability; accordingly, the system is intended to augment, not replace, established clinical judgment. Reconstruction performance depends on adequate image quality and multi-view coverage, and may be affected by lighting variation, occlusion, or patient motion. Future work will focus on prospective multi-center studies, integration of complementary imaging modalities, improved robustness through learning-based reconstruction, and predictive modeling of healing trajectories, as well as deeper integration with electronic health record systems.

In summary, this work demonstrates that advanced computer vision and artificial intelligence techniques can be successfully translated into a clinically meaningful, end-to-end burn management platform. By unifying three-dimensional reconstruction, quantitative analysis, longitudinal monitoring, and clinical documentation within a single system, the proposed approach addresses longstanding challenges in burn assessment and follow-up. The platform has the potential to improve objectivity, reproducibility, and continuity of burn care, supporting more informed clinical decision-making and laying the groundwork for next-generation digital burn management systems.

\bibliographystyle{IEEEtran}

\end{document}